\documentclass[journal]{IEEEtran}

\usepackage[colorlinks,urlcolor=blue,linkcolor=blue,citecolor=blue]{hyperref}

\usepackage{color}
\usepackage{pdflscape}
\usepackage{geometry}
\usepackage{graphicx}
\usepackage{pdflscape}
\usepackage{longtable}
\usepackage{array}
\usepackage{tabularx}
\usepackage{lipsum}
\usepackage{titlesec}
\usepackage{lscape}
\usepackage{adjustbox}
\usepackage{caption}
\usepackage{longtable}
\usepackage{caption}
\usepackage{svg}
\svgsetup{inkscapelatex=false}
\usepackage{cite}
\usepackage{authblk}

\titleformat{\section}[block]{\normalfont\large\bfseries}{\thesection}{1em}{}
\titleformat{\subsection}[block]{\normalfont\normalsize\bfseries}{\thesubsection}{1em}{}
\titleformat{\subsubsection}[block]{\normalfont\normalsize\itshape}{\thesubsubsection}{1em}{}

\renewcommand{\thesection}{\arabic{section}}
\renewcommand{\thesubsection}{\thesection.\arabic{subsection}}
\renewcommand{\thesubsubsection}{\thesubsection.\arabic{subsubsection}}

\begin{document}


\title{On-Device LLMs for SMEs: Challenges and Opportunities}

\author[1]{Jeremy Stephen Gabriel Yee}
\author[1]{Pai Chet Ng}
\author[1]{Zhengkui Wang}
\author[1]{Ian McLoughlin}
\author[2]{Aik Beng Ng}
\author[2]{Simon See}
\affil[1]{Singapore Institute of Technology}
\affil[2]{NVIDIA AI Technology Center}







\maketitle

\begin{abstract} 
This paper presents a systematic review of the infrastructure requirements for deploying Large Language Models (LLMs) on-device within the context of small and medium-sized enterprises (SMEs), focusing on both hardware and software perspectives. From the hardware viewpoint, we discuss the utilization of processing units like GPUs and TPUs, efficient memory and storage solutions, and strategies for effective deployment, addressing the challenges of limited computational resources typical in SME settings. From the software perspective, we explore framework compatibility, operating system optimization, and the use of specialized libraries tailored for resource-constrained environments. The review is structured to first identify the unique challenges faced by SMEs in deploying LLMs on-device, followed by an exploration of the opportunities that both hardware innovations and software adaptations offer to overcome these obstacles. Such a structured review provides practical insights, contributing significantly to the community by enhancing the technological resilience of SMEs in integrating LLMs.
\end{abstract}


\section{Introduction}

The deployment of large language models (LLMs) on consumer and Internet-of-Things (IoT) devices is increasingly crucial across various industries, driven by the need for real-time data processing, enhanced privacy, and reduced reliance on centralized networks \cite{li2024personal}. This transition is particularly pivotal for small and medium-sized enterprises (SMEs), which, despite lacking the extensive infrastructure of larger corporations, require advanced AI capabilities to remain competitive. 
Previous work by O'Leary \cite{10453397} discusses the customization of LLMs for specific enterprise needs with retrieval augmented generation (RAG) while highlighting the limitations of LLMs in terms of operational efficiency and knowledge management in enterprise settings. 
As SMEs adopt smart technologies, the challenge of deploying LLMs in environments where computational power, memory, and energy are limited is becoming a critical area of focus.

Existing work extensively explores the optimization of LLMs through various methods aimed at enhancing model efficiency, refining training processes, and applying compression techniques. These approaches enable LLMs to function effectively on devices with constrained storage and processing power, utilizing methods such as quantization \cite{qin2024accurate}, structured pruning \cite{guo2023compresso}, knowledge distillation \cite{gu2023minillm}, and other innovative approaches to model compression. However, while quantization, model pruning, and knowledge distillation are effective in reducing the computational load and memory requirements, they primarily focus on model-level optimizations without a comprehensive examination of the broader infrastructure challenges associated with deploying these models on edge devices.

For SMEs, understanding how to effectively integrate and optimize LLMs within their more limited hardware and software ecosystems is crucial. These businesses often face unique challenges due to their constrained resources and less specialized technical infrastructure. Despite the pressing need, there has been little systematic exploration of the infrastructure requirements and efficient deployment strategies for LLMs in such contexts. 
Notably, while model compression techniques significantly advance model efficiency, none of these methods extensively examine the implications from both software and hardware perspectives of infrastructure efficiency.

Clearly, for the successful deployment of optimized LLMs in resource-constrained environments, it is essential to consider not only model compression but a holistic view of infrastructure efficiency \cite{bai2024beyond, menghani2023efficient}. The umbrella of infrastructure efficiency encompasses both software and hardware efficiency innovations which are specifically designed for such deployments .

This paper aims to bridge this gap by reviewing the specific challenges and strategies necessary for deploying LLMs in resource-constrained environments typical of SMEs, enabling smaller enterprises to effectively leverage advanced AI models within their limited infrastructural capabilities.

\begin{figure*}[ht]
    \centering
    \includegraphics[width=\linewidth]{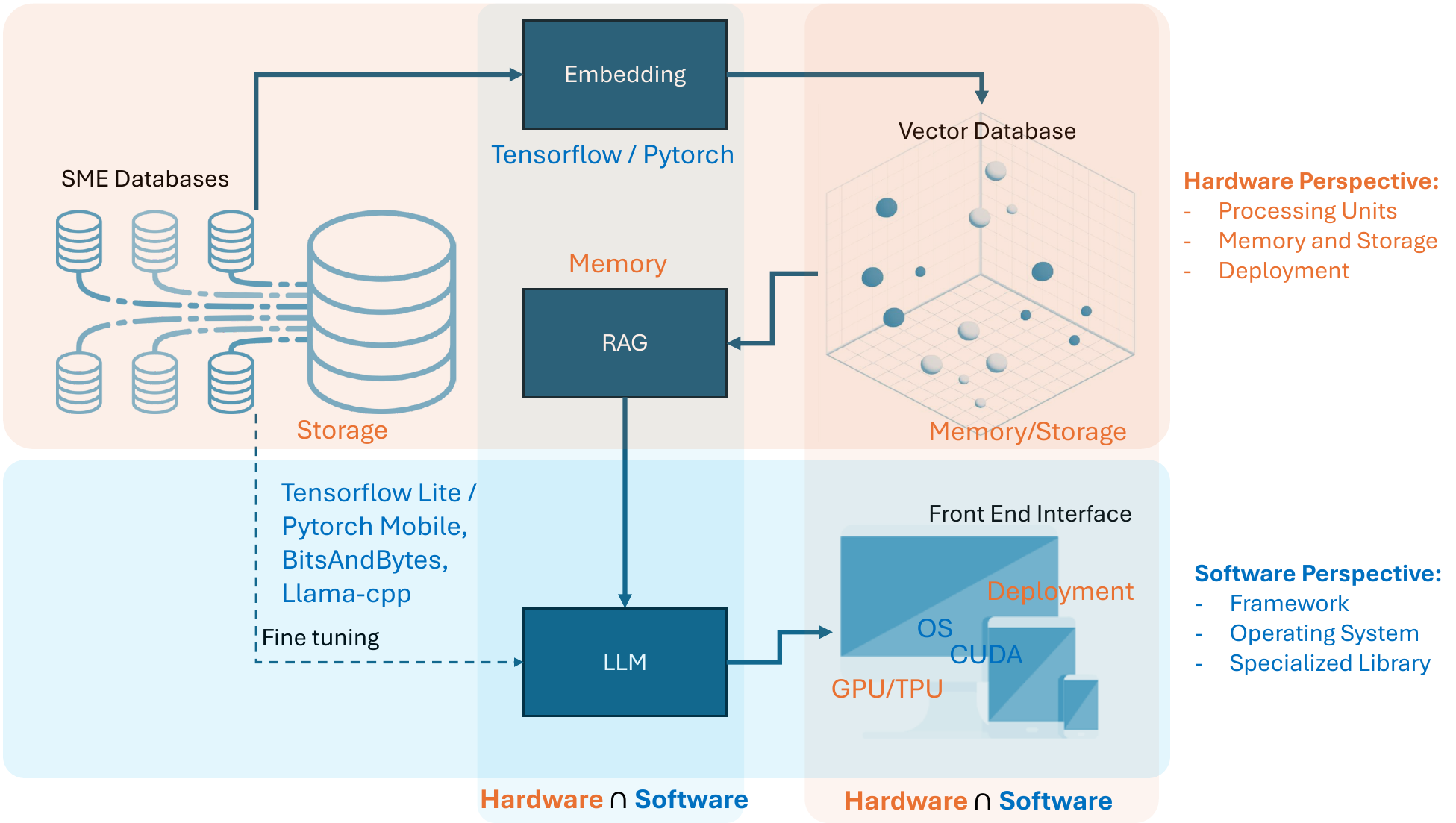}
    \caption[]{Operationalization of LLMs in SMEs Under Resource-Constrained Environments: This diagram illustrates the essential components involved in deploying LLMs within SMEs, focusing on RAG and fine-tuning processes, integrated with local SME databases. Key components for hardware innovations, including processing units and storage solutions, are emphasized in an orange box. Components related to software operations are outlined in a blue box. Areas where hardware and software considerations intersect are depicted in overlapping sections of the boxes. }
    \label{fig:hardware-software}
\end{figure*}

\section{Resource-Constrained Environments}
Resource-constrained environments, characterized by limited computational power, memory, storage, and energy, are common in consumer and IoT devices \cite{10496904}. These environments typically cannot support the heavy demands of high-parameter models during inference due to their limited hardware capabilities.
Figure~\ref{fig:hardware-software} illustrates the key techniques employed to operationalize LLMs effectively within SME settings. Methods such as RAG \cite{10447448} and fine-tuning \cite{10386911}, which utilize local SME databases, are explored to customize LLMs so that to enhance model relevance and performance by leveraging data that is pertinent to the local context.

However, there is a growing trend toward deploying LLMs within these resource-constrained infrastructures in SMEs, driven by the need for smarter, more autonomous applications that prioritize data privacy and reduce dependence on continuous cloud connectivity. As these devices become increasingly embedded with intelligent functionalities for enterprise applications, there is a critical demand for LLMs that are computationally efficient and can operate independently of extensive network infrastructures.
A previous survey by Xu et al. \cite{xu2024survey} focuses on resource-efficient strategies for deploying large foundation models, highlighting the importance of algorithmic and systemic approaches to manage hardware resources effectively. While it provides a comprehensive analysis of model architectures, training algorithms, and system designs to enhance scalability and sustainability, it does not specifically address the unique challenges faced by SMEs nor does it explicitly distinguish between software and hardware perspectives as distinct areas of focus. 
Our work, on the other hand,  explicitly considers the SME context and delineates strategies from both software and hardware perspectives.

As illustrated by the orange and blue boxes  in Figure~\ref{fig:hardware-software}, this paper reviews the specific challenges and opportunities associated with deploying LLMs in resource-constrained environments from two key perspectives:
\begin{itemize}
    \item \textbf{Software Perspective}: This includes considerations for framework compatibility, operating system compatibility, and the utilization of specialized libraries to optimize the performance of LLMs in resource-constrained environments.
    \item \textbf{Hardware Perspective}: This focuses on evaluating processing units (GPUs/TPUs), optimizing memory and storage solutions, and enhancing deployment efficiency to support the operation of LLMs in environments with limited hardware capabilities.
\end{itemize}

\section{Software Framework}
For SMEs, the strategic selection and effective utilization of software frameworks are fundamental to the successful deployment and operation of LLMs within their limited resource environments. The right software frameworks help in minimizing memory footprint, enhancing processing speeds, and ensuring that LLMs can run effectively even on devices with limited computational resources. By focusing on these frameworks, SMEs can not only ensure optimal performance but also maintain competitive advantage in an increasingly AI-driven marketplace.

\subsection{Operating System Compatibility}
For SMEs with diverse technological ecosystems, the choice of operating system (OS) can significantly influence the performance and efficiency of deployed machine learning models, particularly LLMs. Operating systems like Linux \cite{714831} offer a compelling choice for many SMEs due to their open-source nature, which not only reduces costs but also provides robust community support. This support is crucial for accessing a wide range of specialized tools and libraries developed specifically for enhancing machine learning workflows. Moreover, Linux's compatibility with various technologies facilitates the customization and optimization of LLMs to meet specific operational needs, making it a preferred OS in resource-constrained devices.

Recent innovations in operating systems, such as AIOS \cite{mei2024aios} and LLM as a System Service (LLMaSS) \cite{yin2024llm}, are transforming the integration of LLMs within SME settings. AIOS embeds LLMs as core components of the OS, enhancing resource allocation, facilitating agent interaction, and supporting concurrent execution, thereby boosting the system's cognitive capabilities and addressing common bottlenecks in environments with diverse agents. Meanwhile, LLMaSS redefines LLM deployment on mobile devices by maintaining persistent states crucial for personalized interactions, introducing techniques like Tolerance-Aware Compression and IO-Recompute Pipelined Loading to minimize latency and optimize memory under tight constraints.

The adaptability of an OS to support cross-platform functionalities is indispensable for SMEs that utilize mixed technology infrastructures. Ensuring that the frameworks and libraries employed are compatible across different operating systems — including Windows, macOS, and Linux — is essential. This compatibility allows for seamless integration and consistent functionality of LLMs across diverse hardware setups, enhancing the operational flexibility and technological resilience of SMEs. By prioritizing cross-platform support, SMEs can leverage their existing hardware more effectively, ensuring that their investments in LLM technology deliver maximum return and support scalable, future-proof business operations.

\begin{table*}[h]
\centering
\caption{Comparison of GPU and TPU for On-Device LLM Deployment in SMEs}
\resizebox{\textwidth}{!}{
\begin{tabular}{|l|l|l|}
\hline
\textbf{Feature} & \textbf{GPU (Graphics Processing Unit)} & \textbf{TPU (Tensor Processing Unit)} \\ \hline
\textbf{Design Focus} & General-purpose parallel computing & Optimized specifically for tensor operations \\ \hline
\textbf{Performance} & High performance in a wide range of AI tasks & Exceptionally high performance in deep learning tasks \\ \hline
\textbf{Efficiency} & Good performance per watt, varying by model & Superior performance per watt, designed for energy efficiency \\ \hline
\textbf{Flexibility} & Supports a broad range of deep learning and general algorithms & Primarily focused on deep learning models \\ \hline
\textbf{Ecosystem} & Mature ecosystem with extensive support for various frameworks & Limited to specific frameworks optimized for TPUs \\ \hline
\textbf{Cost} & Consumer-grade GPUs are widely available and cost-effective & Generally more expensive and less accessible \\ \hline
\textbf{Integration} & Easier integration with existing systems and software & Requires specific software and infrastructure setup \\ \hline
\textbf{Use Case} & Ideal for SMEs with diverse computational needs & Best for SMEs focused heavily on deep learning deployments \\ \hline
\textbf{Deployment} & Flexible deployment in desktops, servers, and embedded systems & Typically used in cloud environments or specialized setups \\ \hline
\end{tabular}
}
\label{table:gpu_vs_tpu}
\end{table*}

\subsection{Compute Unified Device Architecture}
Compute Unified Device Architecture (CUDA) is a parallel computing platform developed by NVIDIA that enables dramatic performance enhancements by allowing software developers to use a high-level programming language to harness the power of NVIDIA GPUs \cite{nvidia_cuda2024}. For SMEs, adopting CUDA can lead to significantly faster data processing speeds and more efficient machine learning operations compared to traditional CPU-based computing. For SMEs that require intensive computational tasks, the upfront investment in NVIDIA GPUs can be cost-effective over the long run. This is particularly true when compared to recurring costs associated with cloud-based GPU services, where fees are incurred for ongoing usage.

Integrating CUDA into SMEs' computational framework can dramatically enhance the performance of LLMs not only during the training and fine-tuning phases but also throughout the inference stage, which is critical for front-end applications that interact directly with users. Inference tasks, such as processing natural language queries or generating content in real-time, benefit immensely from CUDA’s ability to handle multiple operations concurrently. This reduces latency and improves the responsiveness of AI-driven applications, which is crucial for maintaining a seamless user experience. Moreover, the investment in NVIDIA GPUs and the implementation of CUDA can be a cost-effective solution for SMEs in the long term. By leveraging local processing power, SMEs can avoid the high recurring costs associated with cloud-based GPU services and gain control over their data processing workflows, enhancing both performance and data privacy—an essential consideration for businesses handling sensitive information.

\subsection{Specialized Library}
The choice of a deep learning framework, such as TensorFlow \cite{tensorflow2024} or PyTorch \cite{pytorch2024}, should align with the SME's specific requirements, including compatibility with existing infrastructure, ease of use, and community support. For SMEs interested in deploying machine learning models directly onto consumer-facing hardware such as mobile phones and embedded systems, frameworks like TensorFlow Lite and PyTorch Mobile offer specialized solutions. These frameworks are designed to be lightweight, ensuring that they can operate efficiently on devices with limited computing power and storage capacity. TensorFlow Lite and PyTorch Mobile not only facilitate faster model inference on edge devices but also support a wide range of optimization tools to reduce the model size without compromising performance.

In addition to standard frameworks, specialized libraries play a pivotal role in adapting LLMs for use in diverse technological setups. Libraries like BitsAndBytes \cite{dettmers2023spqr} and Llama-cpp-python \cite{abetlen2024llama} enhance the adaptability of LLMs by focusing on quantization and compression techniques, which are critical for managing memory usage and improving inference speed on limited-capacity devices. BitsAndBytes, for instance, offers k-bit quantization options that significantly reduce memory demands while maintaining performance, making it possible to run large models on standard GPUs. On the other hand, Llama-cpp-python provides a versatile solution for non-CUDA environments by supporting multiple backends such as CPU-only, CUDA, Metal, and OpenCL. This cross-platform capability ensures that SMEs can deploy sophisticated AI models across various hardware platforms, expanding the reach and applicability of their AI-driven initiatives. These tools collectively empower SMEs to navigate the challenges of limited hardware capabilities, ensuring that their investments in AI yield effective, scalable solutions suitable for a wide range of applications.

\section{Hardware Innovations}
For SMEs, tapping into the latest hardware innovations for deploying LLMs on-device presents an opportunity to leverage advanced AI capabilities efficiently. By carefully selecting appropriate GPUs or TPUs, optimizing memory and storage, and utilizing cutting-edge edge computing solutions, SMEs can overcome the typical barriers associated with high-performance AI applications. This strategic approach ensures that even resource-constrained devices can deliver powerful AI functionalities, enhancing business operations and offering competitive advantages in a technology-driven marketplace.

\subsection{High-Performance GPU and TPU}
Graphic Processing Units (GPUs) and Tensor Processing Units (TPUs) represent two distinct paths for accelerating the computational capabilities required for training and deploying LLMs in SME settings, each with unique advantages as highlighted in Table~\ref{table:gpu_vs_tpu}. The choice between GPUs and TPUs for an SME will depend on the specific requirements of their LLM applications—whether the priority is on general versatility and ease of integration offered by GPUs or the specialized, high-efficiency operations provided by TPUs. Both units play critical roles in facilitating the deployment of LLMs, enhancing the technological resilience and operational efficiency of SMEs in integrating advanced AI models.

GPUs are versatile, making them essential for a broad range of computational, including deep learning \cite{mittal2019survey}. Their ability to perform parallel operations, such as matrix multiplications, is crucial for the efficient execution of LLM tasks. For SMEs, the use of consumer-grade GPUs is particularly attractive due to their affordability and ready availability, which allows for significant enhancements in LLM operations without extensive infrastructure investments. The mature ecosystem surrounding GPUs ensures a wealth of libraries and tools are available, facilitating easier integration and optimization of LLM tasks across various platforms.

On the other hand, TPUs are specifically engineered for high-throughput tensor operations, particularly in terms of performance per watt—a critical factor in deployment phases where inference speed and power efficiency are paramount \cite{shahid2020survey}. Designed to optimize operational costs while maximizing performance, TPUs are ideal for SMEs that are heavily focused on deep learning deployments. However, TPUs often require specific software and infrastructure setups, which might limit their use to more specialized applications compared to the more flexible GPUs.

\subsection{Memory and Storage}
Efficient memory management is crucial for running LLMs on devices with limited RAM and storage capacity. Techniques such as model quantization and pruning, reviewed by \cite{xu2024survey}, are also beneficial from a hardware perspective as they reduce the memory footprint of LLMs, making them feasible for deployment on less capable hardware. 
From the hardware perspective, the PagedAttention system \cite{kwon2023efficient} represents a significant advancement in this area, utilizing techniques inspired by virtual memory and paging in operating systems to optimize the use of key-value (KV) cache memory. This approach minimizes waste by preventing fragmentation and reducing redundancy, allowing for more effective batching of LLM requests which is crucial for high-throughput operations. 
For situations where it is reasonable to use multiple LoRA fine-tuned models, the LoRAX framework \cite{zhao2024lora} facilitates the deployment of numerous models on a single GPU, the framework works by focusing on a shared weights of a base model and dynamic adapter loading. The sharing of weights allow for a cost-effective and efficient host of a wide range of specialized LLMs which makes it suitable for a resource-constrained multi-agent environment.

The deployment of LLMs for SMEs necessitates innovative approaches to manage storage constraints effectively. Techniques such as the memory-efficient Differentiable KMeans Clustering (eDKM) \cite{10423861} have shown promising results in compressing LLMs to fit within the limited storage capacities typical of mobile devices. For instance, eDKM successfully compresses a pretrained LLaMA 7B model significantly, reducing its size from 12.6 GB to 2.5 GB, thereby facilitating on-device deployment without compromising performance.   Integrating LLMs with analytical databases such as Redshift or Databricks through user-defined functions allows SMEs to leverage LLM capabilities within their existing data ecosystems \cite{liu2024optimizing}. This integration can process natural language tasks efficiently within analytical workloads, though it also highlights the challenge of managing substantial computational resources. Optimizing LLM inference, as demonstrated in enhancements within Apache Spark, can significantly reduce the latency of database interactions. Such advancements are crucial for SMEs that rely on rapid data processing and real-time analytics, underscoring the need for sophisticated storage management solutions that can handle not only the LLMs themselves but also the vast amounts of data processed and generated by these models.

\subsection{Deployment Efficiency}
Deploying LLMs on edge devices requires careful consideration of the hardware capabilities of these devices. Innovations in edge computing hardware, such as NVIDIA's Jetson series, provide powerful computing capabilities in compact form factors suitable for edge deployment. These devices are engineered to handle AI tasks efficiently, making them ideal for SMEs looking to implement LLMs closer to data sources, thus reducing latency and improving response times in applications such as real-time language translation or decision support systems.

The move towards autonomous edge AI platforms, especially evident with the advent of 6G networks, illustrates the ongoing evolution in how AI models, including LLMs, are adapted to meet the rigorous demands of edge computing scenarios \cite{9606720}. This adaptation is essential for SMEs operating in sectors where rapid data processing and decision-making are critical, yet access to traditional data centre resources is limited. For instance, Dhar et al. \cite{dhar2024empirical} explore the deployment of LLMs on edge devices, identifying key challenges such as insufficient memory and computing resources on traditional edge devices. They provide valuable insights and design guidelines that can significantly aid SMEs in optimizing LLM deployments to enhance operational efficiency and scalability.

Furthermore, the integration of LLMs into mobile and edge computing environments is not just about improving computational efficiency but also enhancing user privacy and system responsiveness. The work by Li et al. \cite{li2024transformer} on Transformer-Lite demonstrates substantial advancements in deploying LLMs on mobile phone GPUs, achieving significant speed improvements in model inference, which is vital for applications requiring immediate feedback. Such developments validate the potential of edge-based LLM deployments to transform SME capabilities, enabling them to leverage powerful AI tools within their limited infrastructural frameworks effectively.

\section{Conclusion}
The exploration of software frameworks and hardware innovations for on-device deployment of large language models (LLMs) reveals significant advancements and opportunities for SMEs. The use of specialized frameworks like TensorFlow Lite and PyTorch Mobile, along with optimization techniques such as dynamic quantization and model pruning, has made it feasible to deploy sophisticated LLMs on edge devices. The incorporation of hardware solutions like GPUs and TPUs has further enhanced the performance, making real-time AI processing accessible in resource-constrained environments. These developments allow SMEs to leverage cutting-edge machine learning capabilities without the need for extensive infrastructure traditionally associated with big tech companies.

Future research opportunities include improving energy efficiency, where innovations in low-power electronics and energy-aware computing could extend device battery life and reduce environmental impact. 
The development of custom hardware solutions like application-specific integrated circuits (ASICs)  could significantly boost processing speeds and cost efficiency. These research areas not only address technical challenges but also promise substantial advancements in LLM deployment across various computational environments.

\section{Acknowledgements}
This work is supported by the SIT-NVIDIA Joint AI Centre

\bibliographystyle{IEEEtran}
\bibliography{ref}

\section{About the Authors}
\textbf{Jeremy Stephen Gabriel Yee} is currently working towards his doctorate with the SIT-NVIDIA Joint AI Centre in Singapore. His research interests include foundational models, multimodal models, and computer vision. He received his master's degree in Scientific Computing and Data Analysis from Durham University. 

\textbf{Pai Chet Ng}  is currently an Assistant Professor in the Infocomm Technolgy Cluster at Singapore Institute of Technology (SIT). Prior to this, she served as a postdoctoral fellow at the University of Toronto and a research associate at the University of Guelph. She earned her Ph.D. from the Hong Kong University of Science and Technology in 2020. Her research interest focuses on Applied AI on consumer devices including mobile and wearable devices for human behaviour analysis with behavioural and physiological signal processing, multimodal biometrics specifically on the area of contactless biometrics, skin analysis with reconstructed hyperspectral image, proximity networking and IoT sensing.  

\textbf{Zhengkui Wang}  is currently an Associate Professor in the Infocomm Technolgy Cluster,  the director of SIT-NVIDIA Joint AI Centre, and the director of the Data Science and AI Lab (DSAIL) at the Singapore Institute of Technology (SIT).  He received his Ph.D. from the National Unversity of Singapore in 2013. His research interests are Machine Learning and Deep Learning, Text Mining, Natural Language Processing, Image Recognition/Annotation/Clustering,  Graph Neural Networks,  Big Data, and Data Warehousing.  His works have been published in more than 60 papers in various prestigious international conferences and journals in these domains, such as AAAI, IJCAI, NuerIPS, SIGMOD,  ICDE, KDD, FSE, IEEE Transactions on Pattern Analysis and Machine Intelligence, IEEE Transactions on Knowledge and Engineering, Bioinformatics etc.

\textbf{Ian McLoughlin}  is currently a  Professor and ICT Cluster Director in Singapore Institute of Technology (Singapore's 5th university) was previously a professor and Head of the School of Computing at the University of Kent (Medway Campus) from 2015-2019, a professor at the University of Science and Technology of China NELSLIP lab from 2012-2015. Before that he spent 10 years at Nanyang Technological University, Singapore and 10 years in the electronics R\&D industry in New Zealand and the UK. Professor McLoughlin became a Chartered Engineer in 1998 and a Fellow of the IET in 2013. He has over 200 papers, 4 books and 13 patents in the fields of speech \& audio, wireless communications and embedded systems, and has steered numerous technical innovations to successful conclusions.

\textbf{Aik Beng Ng} is currently the Senior Regional Manager at NVIDIA AI Technology Center (NVAITC) Asia Pacific South. His research interests include AI innovation, strategic partnerships, and technology ecosystem development. He received his Ph.D. in AI from the Singapore University of Technology and Design (SUTD). He is a member of the National AI Technical Committee. 
He leads a team of young researchers at NVAITC Asia Pacific South, working on advancing AI research and promoting AI through collaborations with academia, industry, and government. With over 20 years of experience, his career spans diverse roles across private and public sectors, including leading innovation teams, engaging global companies, and patenting PC software deployment methodologies. He actively shares insights on AI-powered industry transformations through speaking engagements at events like Big Data \& AI World 23 and TechWeek Singapore. He co-organizes research symposiums and has co-authored award-winning papers. As Co-Director of the NV-SIT Joint Center, he continues to drive AI innovation and shape the future of technology.

\textbf{Simon See} is currently the Solution Architecture and Engineering Director, Chief Solution Architect, and Global Head for NVIDIA AI Technology Center (NVAITC) at NVIDIA Corporation. His research interests include High Performance Computing, Big Data, and Artificial Intelligence. He received his Ph.D. in Electrical Engineering and Numerical Analysis from the University of Salford, UK, and is a Fellow of IEEE. He leads global AI initiatives at NVAITC and holds adjunct professorships at Shanghai Jiao Tong University, Coventry University, Universitas Indonesia, and Newcastle University in Singapore. Additionally, he is a distinguished fellow at Fudan University and an Adjunct Scientist at A*STAR’s Institute of High Performance Computing (IHPC). He has served on the Steering Committee of NSCC’s Supercomputing Asia conference since 2018. He has published over 200 papers, received various awards, and provides consultancy to national research and supercomputing centers. He serves on several advisory boards and committees, including the International Advisory Board of the Institute of Operations Research \& Analytics (IORA) and the Machine Intelligence and Data Analytics Research Center (MIDARC). Before joining NVIDIA, he worked with SGI, DSO National Laboratories, IBM, International Simulation Ltd (UK), Sun Microsystems, and Oracle.

\end{document}